\documentclass[conference]{IEEEtran}
\IEEEoverridecommandlockouts

\usepackage{cite}
\usepackage{amsmath,amssymb,amsfonts}
\usepackage{algorithmic}
\usepackage{graphicx}
\usepackage{textcomp}
\usepackage{multirow}
\usepackage{colortbl}
\usepackage{xcolor}
\usepackage{booktabs}
\usepackage[colorlinks,linkcolor=blue]{hyperref}

\definecolor{green1}{RGB}{32,32,180}
\definecolor{green2}{RGB}{72,72,192}
\definecolor{blue1}{RGB}{242,248,242}
\definecolor{deepgreen}{RGB}{117,189,66}
\definecolor{brown}{RGB}{198,95,16}
\definecolor{lowbrown}{RGB}{242,210,190}
\definecolor{lowblue}{RGB}{170,200,250}

\makeatletter
\newcommand{\linebreakand}{%
  \end{@IEEEauthorhalign}
  \hfill\mbox{}\par
  \mbox{}\hfill\begin{@IEEEauthorhalign}
}
\makeatother

\def\BibTeX{{\rm B\kern-.05em{\sc i\kern-.025em b}\kern-.08em
    T\kern-.1667em\lower.7ex\hbox{E}\kern-.125emX}}
\begin{document}

\title{Seeing is Believing? Enhancing Vision-Language Navigation using Visual Perturbations
}



\author{
\IEEEauthorblockN{Xuesong zhang, Jia Li*\thanks{* Corresponding author}, Yunbo Xu, Zhenzhen Hu, Richang Hong }
\IEEEauthorblockA{\textit{The School of Computer Science and Information Engineering, Hefei University of Technology, Hefei, China} \\
\{xszhang\_hfut, xuyunbocn\}@mail.hfut.edu.cn, jiali@hfut.edu.cn,
\{huzhen.ice, hongrc.hfut\}@gmail.com}
\vspace{-18pt}
}



\maketitle

\begin{abstract}


Autonomous navigation guided by natural language instructions in embodied environments remains a challenge for vision-language navigation (VLN) agents. 
Although recent advancements in learning  diverse and fine-grained visual environmental representations have shown promise, the fragile performance improvements may not conclusively attribute to enhanced visual grounding—a limitation also observed in related vision-language tasks.
In this work, we preliminarily investigate whether advanced VLN models genuinely comprehend the visual content of their environments by introducing varying levels of visual perturbations.
These perturbations include ground-truth depth images, perturbed views and  random noise. 
Surprisingly, we experimentally find that simple branch expansion, even with noisy visual inputs, paradoxically improves the navigational efficacy.
Inspired by these insights, we further present a versatile Multi-Branch Architecture (MBA) designed to delve into the impact of both the branch quantity and visual quality. The proposed MBA extends a base agent into a multi-branch variant, where each branch processes a different visual input. This approach is embarrassingly simple yet agnostic to topology-based VLN agents. Extensive experiments on three VLN benchmarks (R2R, REVERIE, SOON) demonstrate that our method with optimal visual permutations matches or even surpasses state-of-the-art results. The source code is available at \href{https://github.com/HCI-LMC/VLN-MBA-VisualPerturbations}{here}. 


\end{abstract}

\begin{IEEEkeywords}
Vision-language navigation, multimodal representation, visual perturbations.
\end{IEEEkeywords}





\section{Introduction}

Recently, vision-language models powered by deep neural networks have facilitated progress in embodied multimodal tasks, such as embodied question answering (EQA) \cite{EQA_2018_CVPR,wang2024embodiedscan}, object rearrangement \cite{pmlrtang23a-rerange}, and vision-and-language navigation (VLN) \cite{anderson2018R2R} , which have gained attention for their potential in human-machine interaction.
Among them, VLN requires egocentric agents to precisely navigate to a target location and even identify queried objects \cite{qi2020reverie,zhu2021soon, shridhar2020alfred} within a visual environment by diverse natural language instructions. To perform this challenging task successfully, agents not only need to comprehend the semantic relationship between language instructions and visual scenes but also adapt to environmental layouts, enabling them to generalize well in unseen environments.

\begin{figure}
    \centering
    \includegraphics[width=0.95\linewidth]{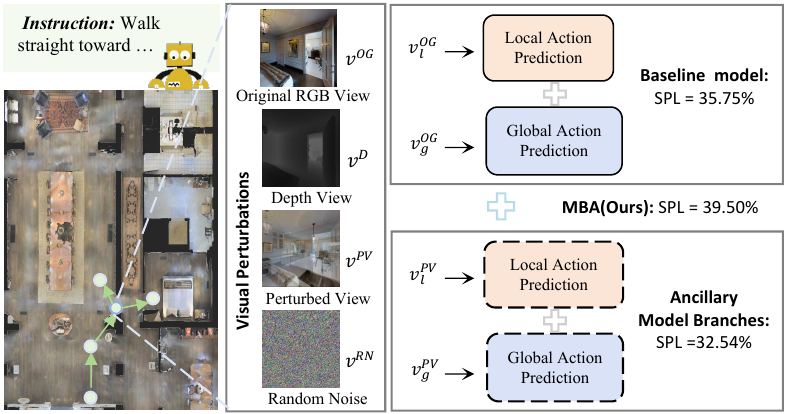}
    \caption{Overview of our multi-branch architecture with different visual input choices for the VLN agent: original RGB (\(v^{OG}\)), Perturbed View (\(v^{PV}\)), corresponding depth images (\(v^{D}\)) and random noise (\(v^{RN}\)). Subscripts \(g\) and \(l\) denote the global and local branches, respectively. SPL metric represents navigational success rate weighted by path length.}
    \label{fig:intro}
    \vspace{-10pt}
\end{figure}



Prevalent navigation prediction strategies can be categorized into Transformer-based local predictions \cite{chen2021hamt, Lin_2024, qiao2022hop} and topology-based global predictions \cite{deng2020gmap1,wang2021gmap2,chen2021topological}. DUET \cite{chen2022duet} marked a significant milestone in VLN tasks by dynamically merging coarse-grained global planning with fine-grained local decision-making by integrating these two strategies.
Based on this dual branch strategy, recent efforts have focused on finer-grained perception of visual environments to further improve navigation metrics, by constructing metric maps \cite{an2023bevbert, hong2023learning} or volumetric representations \cite{liu2024volumetric}, alongside existing topological maps. 
These multifaceted visual representations predominantly rely on RGB observations. However, in related multimodal tasks, studies have revealed that the visual modality is often underexploited\cite{areyou,ilinykh2022look-biasvln,zhu2021diagnosing,shrestha2020negative-biasvqa}, with disproportionate emphasis on language. For example, the notorious  language prior problem in VQA~\cite{zhangVQA} leads to well-trained models providing superficial answers based on question rather than fully engaging with the image information. Ilinykh et al.~\cite{ilinykh2022look-biasvln} show that even when vision is inadequately leveraged in EQA, the model can still derive useful patterns from various visual permutations. Similarly, Zhu et al.~\cite{zhu2021diagnosing} highlight an imbalance in the attention given to vision and text inputs in some VLN benchmarks. 
This prompted us to doubt whether advanced VLN agents are effectively leveraging the visual modality.

In this work, our study serves as a testbed to examine the impact of different levels of visual perturbations on the baseline agent~\cite{chen2022duet}, while also investigating how different types of perturbations can further enhance navigation performance.
The visual input strategies include, in addition to the original RGB view, the following: 1) depth view, which refers to ground-truth depth images providing precise but semantically sparse information; 2) perturbed view, where visual inputs are integrated with incongruent views, introducing a controllable level of visual perturbations; and 3) noisy view, where random noise is infused, resulting in the most disturbed representation.
As shown in Fig. \ref{fig:intro}, when the baseline visual input \((v_g^{OG}, v_l^{OG})\) is replaced by perturbed inputs \((v_g^{PV}, v_l^{PV})\), the agent's primary navigation metric, SPL, does not experience a significant decline.  
Surprisingly, a simple adaptive extension of these two combinations into a four-branch navigation prediction framework dramatically enhances the agent's performance, elevating the SPL metric from 35.75\% to 39.50\%.
These unexpected results offer a fresh insight: current VLN agents are insensitive to visual scene variations and the potential of the visual modality remains underexplored.
To relax this finding to more general VLN settings and further investigate the impact of visual quality and branch expansion, we propose a simplistic but efficacious Multi-Branch Architecture (MBA) which extends the base model by incorporating multiple branches, each of which can receive either identical or diverse visual inputs.
Ultimately, we dynamically combine the outputs of each branch to predict navigation actions based on the optimal visual input permutations. Notably, this approach seamlessly integrates with other topology-based agent models, consistently elevating their navigational generalization capabilities. 

The main contributions of this paper are as follows:
\begin{itemize} 
\item  We experimentally observe that advanced VLN agents, failing to fully capitalize on the visual modality, exhibit performance that is significantly influenced by both visual quality and branch expansion. 
\item We introduce distinct visual inputs and propose a straightforward framework to fuse these features. This allows us to analyze the impact of varying levels of perturbation on VLN models, ultimately identifying the optimal visual input combination to improve generalization. 
\item Experimental results demonstrate that the proposed MBA method effectively enhances agent navigation performance across three VLN benchmarks, significantly outperforming state-of-the-art (SOTA) results on REVERIE \cite{qi2020reverie} and SOON \cite{zhu2021soon}.
\end{itemize} 

\section{Related Work}

\subsection{Vision-Language Navigation}

Vision-language navigation (VLN) research is making significant strides, fueled by its potential applications in household robots and rescue assistants. 
VLN requires an agent to accurately navigate \cite{anderson2018R2R}  or locate target objects\cite{qi2020reverie,zhu2021soon} using natural language instructions in unseen environments. 
Prevalent navigation strategies can be divided into two categories: Transformer-based local prediction and Topology-based global prediction. 
For local prediction, \cite{qiao2022hop,qiao2023hop+,chen2021hamt} employs a hierarchical vision transformer to learn text, current, and historical visual observations, however, such local prediction methods struggle with long-range navigation. In an orthogonal direction,  topological maps \cite{deng2020gmap1,wang2021gmap2,georgakis2022cross} provide global planning but may lack visual details.
DUET \cite{chen2022duet} addresses this by dynamically merging coarse-grained global planning with fine-grained local decision-making via a dual-branch method.
More recently, efforts that further introduced diverse visual environmental representations (such as metric maps \cite{an2023bevbert,wang2023gridmm} and generative goal location images \cite{li2023lad}) to capture more fine-grained visual layouts.
Following these mainstream researches, our study adopts DUET as the baseline agent, given its integration of both local and global navigation categories.

\subsection{The Role of Vision in Visual-Language Tasks.}



Numerous vision-language tasks have been shown to underutilize the visual modality, leading models to disregard or hallucinate visual content. For example, many VQA models \cite{shrestha2020negative-biasvqa,zhangVQA} often rely on superficial correlations between question types and ground truth answers, rather than adequately considering the visual information. 
The EQA model also exhibits a tendency to hallucinate and disregard visual input~\cite{ilinykh2022look-biasvln}. 
Similar phenomenons have also been observed in image captioning \cite{imagecaption1} and even large-scale vision-language models\cite{imagecaption2}. 

While previous studies~\cite{ilinykh2022look-biasvln,wu2024noiseboostalleviatinghallucinationnoise} addressed the issue of noisy visual perturbation in multimodal tasks, their application to VLN remains underexplored. Recent works in the VLN domain~\cite{zhang2021diagnosing,GOAT} have demonstrated that even incorrect visual inputs can lead to good performance due to overfitting to dataset biases, ultimately resulting in poor generalization to unseen scenarios. Efforts such as~\cite{areyou,ilinykh2022look-biasvln} have analyzed this problem by removing or replacing visual inputs, aiming to enable models to make accurate predictions by grounding on visual content.
Therefore, in this work, we focus on exploring the role of the visual modality in VLN tasks rather than language instructions quality \cite{zhu2023does}. Inspired by probe systems with perturbed vision in EQA\cite{ilinykh2022look-biasvln},  we introduce distinct visual perturbations to diagnose how different visual content impacts agent performance.

\section{Methodology}

\subsection{VLN Problem Formulation}

\textit{1) Notation:}
For the discrete indoor VLN tasks, such as REVERIE, the simulation environment \cite{anderson2018R2R} is represented as an undirected graph $\mathcal{G}=\{\mathcal{N},\mathcal{E}\}$, where $\mathcal{N}=\{n_i\}_{i=1}^K$ denotes $K$ navigable nodes, and $\mathcal{E}$ denotes the connectivity edges. At the start of each episode, the agent is initialized on a random node. At each time step $t$, the egocentric agent traverses the graph \(\mathcal{G}\)  based on the instruction $\mathcal{W}$. At each navigable node, the agent can receive RGB panorama \(V_{t}=\{v_{t,i}\}_{i=1}^{n=36}\) composed of surrounding view $v_{t,i}$, attached  corresponding direction information \(E_{t}\). Each panorama also consists of $m$ objects \(O_{t}=\{o_{t,i}\}_{i=1}^{m}\) and contains $d$ candidate navigation actions (or nodes)  \(A_t=\{a_{t, i}\}_{i=0}^d\), where $m$ and $d$ are variable at different navigable nodes.  Based on the instruction $\mathcal{W}$ and visited visual observations, the agent predicts the next node from $A_t$ or stop at the current location $a_{t,0}$ as next action. The agent predicts objects only when it decides to stop.

\textit{2) Baseline Agents:}
We adapt DUET \cite{chen2022duet} as a baseline and first go over its local and global action prediction modules during navigation process. 
At each timestep, the agent can access information about the current navigable nodes, including RGB panoramic images $V_{t}$, objects $O_{t}$, and location information $E_{t}$. It employs a cross-modality LXMERT \cite{lxmert} model to encode these information alongside the instructions $\mathcal{W}$ in a fine-grained manner for local action prediction. The \textbf{local} branch just predicts candidate ghost nodes adjacent to the current location.
As the timestep increases, the agent retains information about previously visited and glimpsed navigable nodes, gradually constructing a topological map with the connectivity edges $\mathcal{E}$. The topological map is combined with instructions to perform coarse-grained cross-modal encoding for global action predictions. The \textbf{global} branch enable the agent to backtrack to previously visited nodes.
Finally, the agent can dynamically fuse the dual-branch prediction. At the end of the navigation, the agent also need to selects the bounding box of the specific object in REVERIE and SOON.

\begin{figure}[t]
    \centering
    \includegraphics[width=0.93\linewidth]{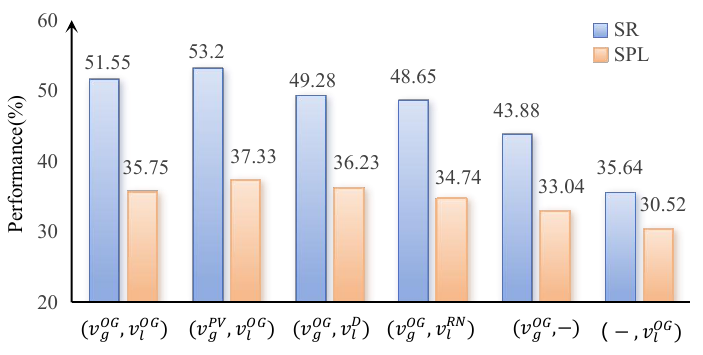}
    \caption{Success Rate (SR) and SPL of the baseline agent with different visual inputs on the REVERIE val-unseen split.}
    \label{fig:find}
    \vspace{-10pt}
\end{figure}


\subsection{The Effect of Visual Quality and Branch Expansion in VLN}

As mentioned above, previous work on vision-language tasks has indicated potential limitations in effective utilization of visual information.  To investigate whether a similar phenomenon exists in VLN, we introduced three distinct visual information (as shown in Fig.~\ref{fig:intro}) to replace the original visual inputs and we also conducted an exploratory analysis.


The results on DUET \cite{chen2022duet} are present in Fig.~\ref{fig:find} preliminary revealed fresh insights into the influence of visual input quality and branch quantity on agent navigation:
\textit{1):} Contrary to expectations, introducing noise to one of the visual input branches did not necessarily degrade performance. Moderately noisy inputs can, in some cases, positively influence navigation performance. For instance, compared to the baseline configuration $(v^{OG}_g, v^{OG}_l)$, which achieved SR and SPL scores of (51.55, 35.75), configurations with noisy inputs, such as $(v^{PV}_g, v^{OG}_l)$ (SR: 53.20, SPL: 37.33) and $(v^{OG}_g, v^{D}_l)$ (SPL: 36.23), even exhibited slight improvements in either SR or SPL. This suggests that the agent demonstrates a degree of robustness to variations in one visual branch when the other branch receives accurate information. 
\textit{2):} The dual-branch architecture of DUET consistently outperformed single-branch counterparts under conditions of varied visual input. For instance, when the local prediction branch was replaced with entirely random noise, the dual-branch configuration $(v^{OG}_g, v^{RN}_l)$ achieved SR and SPL scores of (48.65, 34.74), significantly outperforming single-branch configurations such as $(v^{OG}_g, -)$ (SR: 43.88, SPL: 33.04) and $(-, v^{OG}_l)$ (SR: 35.64, SPL: 30.52). This compellingly demonstrates the superior robustness of dual-branch architectures compared to single-branch (local or global) models when faced with perturbations or noise in the visual input.

These observations highlight the complex interplay between visual input quality and branch expansion in VLN and suggest promising avenues for developing more robust models. To further disentangle the relative contributions of these factors, we will investigate the effects of varying degrees of visual perturbation and increasing the number of visual input branches through a multi-branch architecture on agent performance.

\begin{figure*}[ht]
\centering
\setlength{\abovecaptionskip}{0.1cm}
\includegraphics[width = 0.95\linewidth]{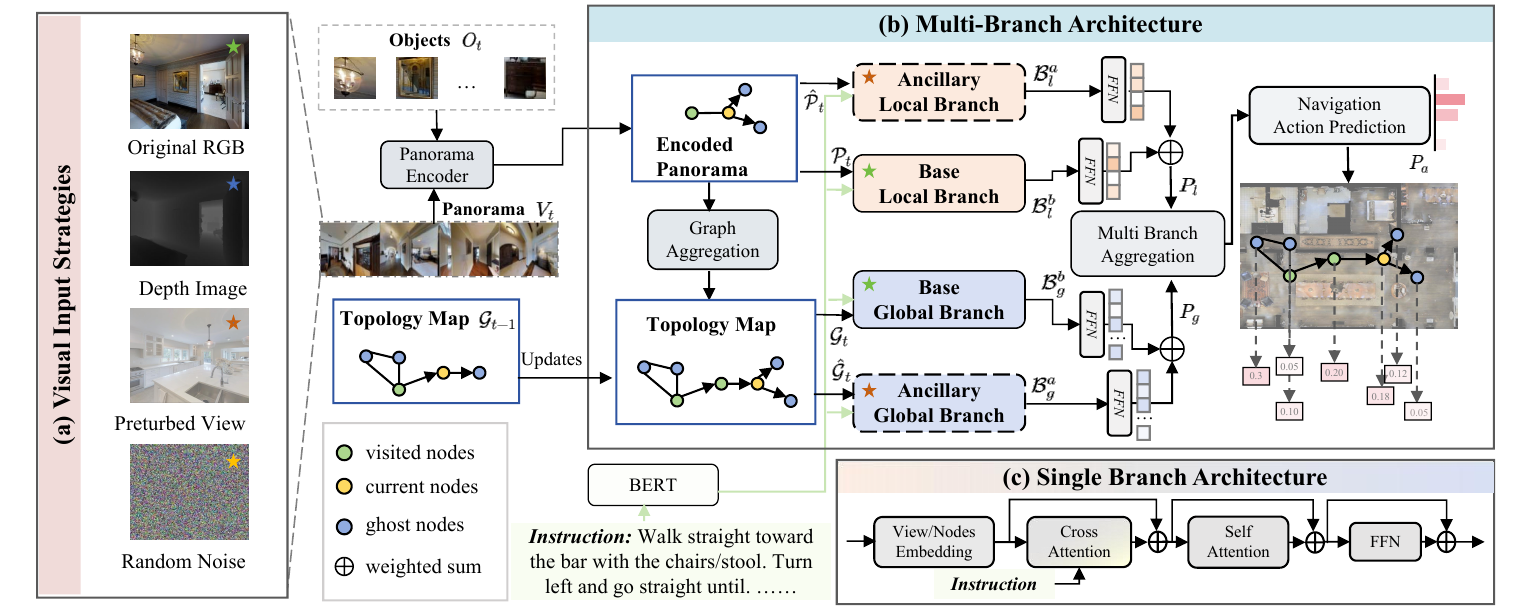}
\caption{Overview of the proposed method, encompassing three components:
(a) presents the visual input strategies, (b) illustrates the pipeline of the Multi-Branch Architecture (MBA) with the optimal visual input combination (labeled by \textcolor{deepgreen}{$\bigstar$} and \textcolor{brown}{$\bigstar$}), and (c) displays the the single-branch architecture, elucidating the internal structure of \textcolor{lowbrown}{local} and \textcolor{lowblue}{global} branch. $\mathcal{\hat{P}}$ and $\mathcal{\hat{G}}$ represent panorama and topological map with perturbed visual inputs.
}
\label{figure: 1}
\vspace{-7pt}
\end{figure*}

\subsection{Visual Perturbations Strategies}

To investigate whether the agent can capture essential information from varying levels of perturbation to enhance navigation performance, here, we formulaically introduce three distinct visual input strategies. 
Specifically, we incorporate ground-truth depth images as a comparative modality, validating that accurate visual information is crucial for effective agent navigation. Drawing inspiration from probe systems with perturbed vision in EQA \cite{ilinykh2022look-biasvln}, furthermore, two distinct visual perturbations are introduced to assess the impact of varying levels of noisy visual content on agent performance and to evaluate the robustness of navigation under challenging visual conditions.
Notably, we only analyze the visual input of each environment view during training, without altering the original instructions $\mathcal{W}$, object features \(O_{t}\) and position information.

\textbf{\textit{Original Visual Inputs.}}
Previous topology-based VLN methods leverage a ResNet-152 \cite{he2016resnet} or ViT-B/16 model \cite{dosovitskiy2020vit} pre-trained for visual feature extraction from each view. However, recent researchs \cite{wang2023dsrg, GOAT, lin2023actional} suggest that CLIP features \cite{radford2021clip} outperform pre-trained ViT features in VLN tasks. Therefore, we adopt CLIP-ViT-L/14@336px to extract RGB image features as the original visual input $v^{OG}$. 

\textbf{\textit{Depth Image.}}
We leverage ground-truth depth images to investigate whether less information-dense yet precise visual inputs can enhance agent navigation.
Specifically, we transform the Matterport Simulator environment into depth views and extract depth features $v^D$ using ResNet-152 \cite{he2016resnet}. To reconcile the dimensional mismatch between depth image features $v^D$ and RGB image features $v^{OG}$, we utilize a feed-forward network (FFN) to project depth images $v^D$ into a 768-dimension space during training.

\textbf{\textit{Perturbed View.}}
FDA \cite{he2024fda} leverages partially low-frequency information from interference images to augment visual inputs. Inspired by this finding, we posit that introducing controlled visual interference in spatial domain could also improve navigation performance. We generate perturbed view by integrating incongruent views $v^{IV}$ and $v^{OG}$, thereby preserving partial correct information: $v^{PV} = (1-\gamma )* v^{OG} + \gamma * v^{IV}$, where $\gamma$ represents a hyperparameter that controls the degree of perturbations.

\textbf{\textit{Random Noise.}}
To conduct a comprehensive analysis and comparison, completely random noisy images are introduced to investigate whether VLN agents can capture any general pattern and commonsense from random noise $v^{RN} \in \mathbb{R}^{768}$, representing the most distorted visual input, devoid of any meaningful content: $v^{RN} = [c_{1},\cdots,c_{768}]$, in which each element $c \in [0, 1]$ represents a random pixel value.

\subsection{Multi Branch Architecture}

To incorporate and fuse the multiple visual inputs described above, we propose a simple yet intriguing approach, termed the \textbf{M}ulti-\textbf{B}ranch \textbf{A}rchitecture (MBA) based on the 
baseline model  DUET \cite{chen2022duet}. 
The pipeline of our proposed framework is shown in Fig.\ref{figure: 1}. 
For a more comprehensive understanding of specific modules, such as the panorama encoder and dynamic fusion, we direct readers to the original paper\cite{chen2022duet}.

\subsubsection{Single Branch Prediction}

At each time step, the agent accesses the encoded panorama $\mathcal{P}_{t}$ of current navigable nodes and predicts adjacent candidate nodes using the local branch. 
As time steps increase, the agent aggregates each visited node graph $\mathcal{G}_{t-1}$ and its adjacent views by average pooling to progressively build a topological map $\mathcal{G}_{t}$. 
It utilizes a global branch to backtrack to earlier visited nodes.
Each single branch, whether global or local, adopts an identical architecture, as illustrated in Fig. \ref{figure: 1}(c).

\subsubsection{Multi Branch Aggregation}
To trade-off and aggregate predictions from each branch, we calculate the learnable weight for each branch, with the 0-$th$ dimension of the different branches serving as the input: $\lambda_l^a,\lambda_g^a,\lambda_l^b,\lambda_g^b = \text{Softmax}(\text{Sigmoid}(\text{FFN}([\mathcal{B}_{l_0}^{a}; \mathcal{B}_{g_0}^{a}; \mathcal{B}_{l_0}^{b}; \mathcal{B}_{g_0}^{b}])))$, where $\mathcal{B}$ represents the last hidden layer of each branch. Then, we predict the navigation scores of each branch using a two-layer feed-forward network: $P = \text{FFN}(\mathcal{B})$.
Ultimately, we dynamically fuse \cite{chen2022duet} predicted distribution of each branch for next action prediction: $P_a=\text{DynamicFuse}(P_l, P_g)$, where $P_l= \lambda_l^a P_l^a + \lambda_l^b P_l^b$ and $P_g = \lambda_g^a P_g^a + \lambda_g^b P_g^b$. For convenience, we have omitted time subscript $t$.


\subsubsection{Training and Inference}
Our pre-training phase followed~\cite{chen2022duet,li2023lad}, utilizing auxiliary tasks such as masked language modeling (MLM), masked region classification (MRC), and single-step action prediction (SAP).
We initialize all fine-tuning models in experiments using same pre-trained baseline with CLIP visual features.
During fine-tuning, the agent is trained to predict both the next action $a^{gt}$ and the final object $o^{gt}$ under cross-entropy (CE) loss supervision. 
Inspired by the DAgger \cite{Dagger} algorithm, the agent also needs to predict a navigable node $a^*$ that is closest to the target destination from the current node:
{\small
\begin{equation}
\begin{split}
    \mathcal{L} = \mu \mathcal{L}_{CE}(P_a, a^{gt}) + \mathcal{L}_{CE}(P_a, a^*) + \mathcal{L}_{CE}(P_o, o^{gt})
\end{split}
    \label{equation: ce-action}
\end{equation}
} where $\mathcal{L}_{CE}$ stands for cross-entropy loss, $P_o$ denotes a prediction score for each object in $O_t$ of the current node.

During inference, agent calculate the shortest path from the current node to the predicted node based on the map. Otherwise, the agent remains at the current location. The agent predicts objects only when the agent decides to stop.

\section{Experiments}
\label{experimentSec}

\subsection{Experiment Settings}

\noindent \textbf{Datasets.}
We primarily validate and analyze our approach on the REVERIE dataset, while also comparing its performance against existing state-of-the-art VLN models across widely used R2R, and SOON datasets. Each dataset presents unique challenges: R2R \cite{anderson2018R2R} requires the agent to navigate to the correct location based solely on detailed instructions; REVERIE \cite{qi2020reverie} demands that the agent selects the correct bounding box from a set of pre-defined object bounding boxes; and SOON \cite{zhu2021soon} requires agents to generate candidate bounding boxes using an object detector, as predefined boxes are absent.

\noindent \textbf{Metrics.} 
We evaluate agents navigation performance using standard metrics. Trajectory Length (TL): Average path length in meters.
Success Rate (SR): Proportion of paths where the agent reaches within 3 meters of the target location.
Navigation Error (NE): Average final distance (meters) between the agent and the target.
Success Rate weighted by Path Length (SPL): Prioritizes success rate for shorter paths.
For REVERIE and SOON tasks, which involve object identification, we also evaluate object grounding metrics: Remote Grounding Success (RGS): Proportion of instructions where the agent correctly identifies the target object.
RGS weighted by Path Length (RGSPL): Prioritizes correct object identification for shorter paths. 
Given that \textbf{SPL} effectively balances navigation success rate and trajectory length, we use it as the primary metric. The SPL formula is as follows: $\text{SPL} = \text{SR} \cdot \frac{d_{gt}}{max(TL, d_{gt})}$, where $d_{gt}$ denotes ground truth distance.

\noindent \textbf{Details.}
The hyperparameter $\gamma$ and $\mu$ are set to a default value of 0.5 and 0.2.
We used batch sizes of 4, 8, and 2 for R2R, REVERIE, and SOON for fine-tuning, respectively.
Unless otherwise specified, we present the DUET experimental results using CLIP as the visual encoder in this paper, whereas the original paper (labeled by DUET$^*$) employed its ViT counterpart.
Additionally, we minimized changes to the baseline~\cite{chen2022duet} settings and used no additional environment annotations.
All experiments were conducted on a single NVIDIA GeForce RTX 4090 with 25k iterations.

\subsection{Ablation Study}
\label{sec: Effect-of-MBA-with-visual-perturbations}

\begin{figure}[t]
\centering
\setlength{\abovecaptionskip}{0.1cm}
\includegraphics[width = \linewidth]{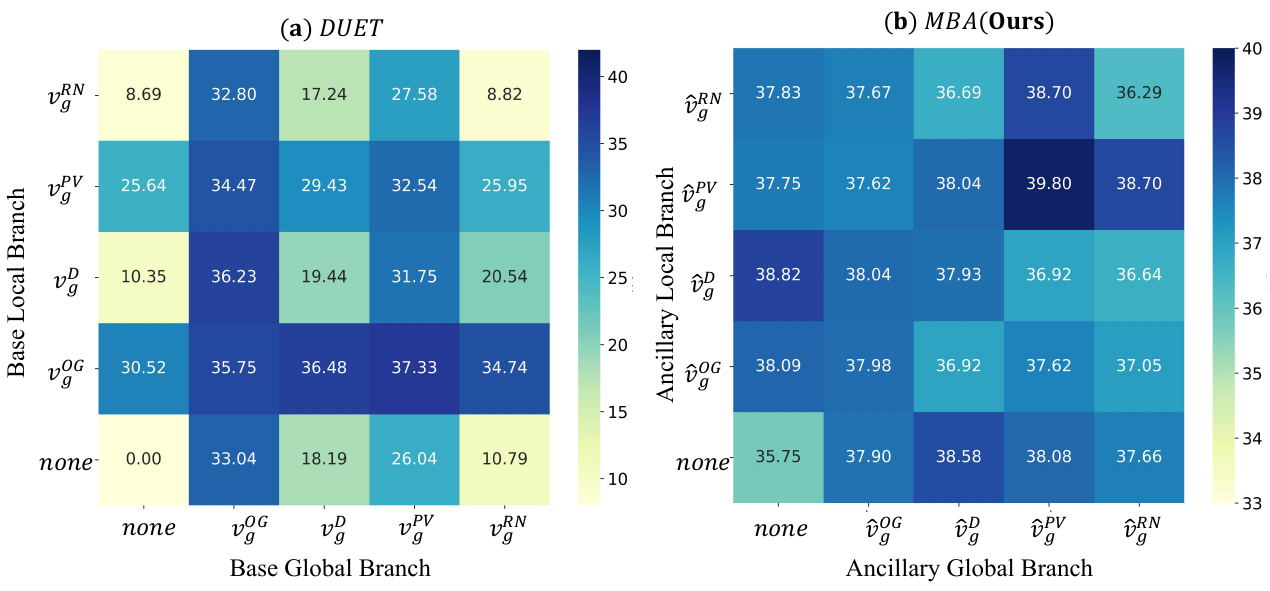}
\caption{The heatmaps show the impact of each branches with diverse visual inputs on the val-unseen split of the REVERIE, with color intensity reflecting the SPL(\%). (a) SPL in DUET for different visual combinations to the base global and local branches. (b) SPL in our MBA model based on DUET for different visual combinations to the ancillary branches. }
\label{figure: 2}
\vspace{-8pt}
\end{figure}

To visually analyze the impact of visual input strategies on navigation performance, we constructed heatmaps based on DUET and our proposed MBA method in Fig.\ref{figure: 2}. Among them, $(v^{OG}_g, v^{OG}_l)$ in Fig.\ref{figure: 2}(a) and $(none, none)$ in Fig.\ref{figure: 2}(b) both represent the DUET baseline \cite{chen2022duet} configuration. In Fig.\ref{figure: 2}(b), leftmost column and bottommost row indicate the three-branch architecture without ancillary single branches, and the remaining cells showcase the results with four-branch architecture.
For convenience, we denote the visual input to the base branch as $v$ and the visual input to the ancillary branch as $\hat{v}$.


\textbf{The Effect of Visual Input Strategies.}
As can be observed in Fig. \ref{figure: 2}(a), disparate visual inputs yield varying performance.  On DUET, both the original visual input and the perturbed input notably outperform the depth view and random noise, suggests that high-quality RGB-based visual cues are essential for effective performance.
Further analysis in Fig. \ref{figure: 2}(b) reveals that depth views performance is enhanced by the ancillary branch, regardless of whether it is local (38.82\%) or global (38.58\%). Moreover, employing two ancillary branches with perturbed views ($\hat{v}^{PV},\hat{v}^{PV}$) led to the best result (39.80\%), even random noise can counterintuitively improve navigation.
This finding suggests that certain noise may act as a regularizer, promoting robust learning and preventing overfitting to specific visual patterns in the training data, thereby improving generalization to unseen environments.


\textbf{The Effect of Branch Expansion.}
Enhancing model capacity with additional branches enables the agent to learn more nuanced and robust representations.
Specifically, compared to the baseline $(none,none)$ 's 35.75\%, even random noise can boost performance to 36.92\% with ancillary branches. The optimal configuration displayed in Fig.\ref{figure: 1}, $(\hat{v}^{PV},\hat{v}^{PV})$, achieves a SPL of 39.80\%.
However, it is important to note that the addition of more branches does not necessarily guarantee better results. For instance, $(\hat{v}^{OG},\hat{v}^{OG})$ achieved a SPL of 37.66\%, lower than $(\hat{v}^{OG},none)$ at 38.09\% and $(none,\hat{v}^{OG})$ at 37.9\%.
This highlights that adding extra branches without careful input selection can lead to over-reliance on the original RGB observations, whereas dynamically adjusting branch weights improves robustness and adaptability in unseen navigation environments.


\begin{table}[t]
\centering
\caption{
SPL Performance (\%) of key components (MBA method, visual perturbations and visual encoder) on REVERIE.
}
\resizebox{0.8\columnwidth}{!}{
\begin{tabular}{cc|c|cc}
\toprule
  MBA   & VP   & Encoder & Val-seen & Val-unseen \\ \midrule
  -   & -     & ViT  & 63.94&            33.73\\
  - & -   & CLIP & 68.03&            35.75\\
  \checkmark & -   &  CLIP & 64.05&     37.66\\
  \checkmark & \checkmark & CLIP & 68.63&  39.80\\ \bottomrule
\end{tabular}}
\label{tab:aba}
\vspace{-7pt}
\end{table}

\textbf{Ablation of MBA with Optimal Visual Permutations.} In the Table \ref{tab:aba}, we present an ablation study of multi-branch architecture with best visual permutations $(\hat{v}^{PV},\hat{v}^{PV})$ on REVERIE val-seen and val-unseen splits. The experimental results presented in the first and second rows that CLIP features are more suitable for visual-language tasks compared to the original ViT features. As shown in row 3, when merely increasing model branches, the model does not exhibit overfitting to seen environments; rather, there is an improvement in SPL within unseen scenarios. Additionally, the result in row 4 indicates that incorporating optimal visual permutations further enhances the model superior adaptability in both seen and unseen environments.
In summary, increasing the model branches to enhances its capacity and introducing diverse visual perturbations to capture essential, generalizable patterns. These factors work synergistically to facilitate agent to adapt to previously unseen navigation scenarios.

\begin{table*}[]
\centering
\caption{Comparison with the state of the art on REVERIE. \textbf{\underline{Bold}} and \textbf{Bold}  highlight the best and runner-up performance in each column, while \colorbox{gray!10}{gray} and \colorbox{blue1}{green} underscore our method row and SPL column. $\uparrow$ indicates better performance with higher values.}
\begin{tabular}{l|ccccc|ccccc}
\toprule
\multirow{2}{*}{ \ \ \ Methods}  &\multicolumn{5}{c}{Val-unseen}&\multicolumn{5}{c}{Test-unseen}\\ & 
\multicolumn{1}{c}{TL} &
\multicolumn{1}{c}{SR$\uparrow$} & 
\multicolumn{1}{c}{\textbf{SPL$\uparrow$} } &
\multicolumn{1}{c}{RGS$\uparrow$} & 
\multicolumn{1}{c|}{RGSPL $\uparrow$}
 &TL  & SR $\uparrow$                & \textbf{SPL$\uparrow$}                & RGS$\uparrow$               & RGSPL$\uparrow$          \\      \midrule
 
 \ \ \ HOP+   \cite{qiao2023hop+} (TPAMI'23)    &14.57  & 36.07                & \cellcolor{blue1}31.13                & 22.49                & 19.33         &15.17        & 33.82                & \cellcolor{blue1}28.24                & 20.20                & 16.86                \\
 \ \ \ DSRG  \cite{wang2023dsrg} (IJCAI'23) & -       & 47.83                & \cellcolor{blue1}34.02                & 32.69                & 23.37    & -             & 54.04               & \cellcolor{blue1}37.09                & 32.49                & 22.18                \\
 \ \ \ BEVBert  \cite{an2023bevbert} (ICCV'23)& -     & 51.78                & \cellcolor{blue1}36.47                & 34.71                & 24.44  & -               & 52.81                & \cellcolor{blue1}36.41                & 32.06                & 22.09                \\
 \ \ \ GridMM  \cite{wang2023gridmm} (ICCV'23) & 23.20    & 51.37                  & \cellcolor{blue1}36.47                & 34.57                & 24.56       &19.97          & 53.13                & \cellcolor{blue1}36.60                & \textbf{34.87}                & \textbf{23.45}                \\
 \ \ \ LAD  \cite{li2023lad} (AAAI'23)   &  26.39     & \textbf{\underline{57.00}}             & \cellcolor{blue1}37.92                &\textbf{\underline{37.80}}                & \textbf{24.59}           & 25.87     &\textbf{56.53}             & \cellcolor{blue1}37.80   &\underline{\textbf{35.31}}             & 23.38    \\
 \ \ \ FDA \cite{he2024fda} (NeurIPS'24)    &  19.04        & 47.57                & \cellcolor{blue1}35.90                & 32.06                & 24.31       &17.30          & 49.62                & \cellcolor{blue1}36.45                & 30.34                & 22.08                \\
 \ \ \ CONSOLE \cite{Lin_2024}  (TPAMI'24) &- &50.07 &\cellcolor{blue1}34.40 &34.05 &23.33 &- &55.13&\cellcolor{blue1}37.13&33.18&22.25          \\      
 \ \ \ KERM \cite{Li_2024} (CVPR‘23) &21.85&49.02& \cellcolor{blue1}34.83&33.97&24.14&18.38&52.26& \cellcolor{blue1}37.46& 32.69&23.15\\
 \ \ \ VER \cite{liu2024volumetric}(CVPR'24) &23.03  &\textbf{55.98}  &\cellcolor{blue1}\textbf{39.66} &33.71  &23.70  &24.74 &\underline{\textbf{56.82}}  &\cellcolor{blue1}38.76&33.88&23.19 \\
\midrule
 \ \ \ DUET$^*$ \cite{chen2022duet} (CVPR'22) &22.11     & 46.98                & \cellcolor{blue1}33.73                & 32.15                & 23.03          & 21.30      & 52.51                & \cellcolor{blue1}36.06                & 31.88                & 22.06                \\
\rowcolor{gray!15}  \ \ \ \ \ \ \ \textbf{+MBA (Ours)} &21.73   &52.29  &  \cellcolor{blue1}\textbf{\underline{39.80}}& \textbf{35.81} & \textbf{\underline{27.44}}   & 18.11   &51.97 &\cellcolor{blue1}\underline{\textbf{39.53}} &33.57 &\textbf{\underline{24.94}} \\ 

 \bottomrule
\end{tabular}
\label{tab:reverie-sota}

\end{table*}

\begin{table*}[t]
\centering
\caption{Comparison with state of the art on the R2R dataset. $\downarrow$ indicates better performance with lower values.}
\begin{tabular}{l|cccc|cccc|cccc}
\toprule
\multirow{2}{*}{\ \ \ \ Methods} &    \multicolumn{4}{c|}{Val-seen}&\multicolumn{4}{c|}{Val-unseen} & \multicolumn{4}{c}{Test-unseen} \\
  & TL   &SR $\uparrow$ &\textbf{SPL}$\uparrow$& NE $\downarrow $&TL  & SR $\uparrow$  & \textbf{SPL}$\uparrow$  & NE $\downarrow$  & TL  & SR$\uparrow$  & \textbf{SPL}$\uparrow$  & NE$\downarrow$ \\ \midrule
\ \ \ \ HOP \cite{qiao2022hop} (CVPR‘22) & 11.26   &75 &\cellcolor{blue1}70&2.72&12.27  & 64   & \cellcolor{blue1}57   & 3.80 & 12.68 &  64  &  \cellcolor{blue1}59  &  3.83 \\
\ \ \ \ HAMT \cite{chen2021hamt} (NeurIPS'21)  &11.15    &76 &\cellcolor{blue1}72&2.51&11.46  & 66   & \cellcolor{blue1}61   & 3.29 &  12.27 &  65  &  \cellcolor{blue1}60  &  3.93 \\
\ \ \ \ DSRG \cite{wang2023dsrg} (IJCAI‘23)   &-    &\textbf{\underline{81}} & \cellcolor{blue1}\textbf{\underline{76}}&2.23&-     & 73   & \cellcolor{blue1}62  &\textbf{\underline{3.00}}  &  -     & \textbf{72}  &  \cellcolor{blue1}61  &  3.33 \\
\ \ \ \ KERM \cite{Li_2024} (CVPR‘23) &12.16&\textbf{80}&\cellcolor{blue1}74&\textbf{2.19}&13.54&72&\cellcolor{blue1}61&3.22&14.60&70&\cellcolor{blue1}59 &3.61\\
\ \ \ \ FDA \cite{he2024fda} (NeurIPS'24) &-    &- &\cellcolor{blue1}-&-&13.68  & 72   & \cellcolor{blue1}\textbf{\underline{64}}   & 3.41 & 14.76 &  69  & \cellcolor{blue1}\textbf{62}  &  3.41 \\
\ \ \ \ CONSOLE \cite{Lin_2024} (TPAMI‘24) &12.74 &79 &\cellcolor{blue1}73 &\underline{\textbf{2.17}} &13.59 &\textbf{73}  &\cellcolor{blue1}63 &3.00 &14.31  &72 &\cellcolor{blue1}61 &\underline{\textbf{3.30}} \\
\midrule
\ \ \ \ DUET$^*$ \cite{chen2022duet} (CVPR’22)& 12.90   &75 &\cellcolor{blue1}69 & 2.63 &12.56  & 72 & \cellcolor{blue1}63 & 3.17 & 14.39  &  71  &  \cellcolor{blue1}61  &  3.31 \\
\rowcolor{gray!15} \ \ \ \ \ \ \ \ \textbf{+MBA (Ours)}  & 12.37   &79&\cellcolor{blue1}\textbf{74} &2.34  &13.39 & \textbf{\underline{73}} & \cellcolor{blue1}\textbf{63} & \textbf{3.05} & 14.26  & \textbf{\underline{72}}   & \cellcolor{blue1}\textbf{\underline{62}}  &  \textbf{3.31} \\
\midrule
\end{tabular}
\label{tab:r2r-sota}
\vspace{-7pt}
\end{table*}

\begin{table}[t]
\centering
\caption{Comparison with the state of the art on the SOON dataset.}
\resizebox{\columnwidth}{!}{
\begin{tabular}{l|cccc}
\toprule
\multirow{1}{*}{Methods} 
                         & TL     & SR $\uparrow$     & \textbf{SPL} $\uparrow$     & RGSPL$\uparrow$   \\ \midrule
                         \multicolumn{5}{c}{Val-unseen} \\ \midrule
GBE  \cite{zhu2021soon} (CVPR2021)                   & 28.96    & 19.52   & \cellcolor{blue1}13.34   & 1.16    \\
GridMM \cite{wang2023gridmm}  (ICCV'23)                & 38.92    & 37.46   & \cellcolor{blue1}24.81   & 3.91    \\
GOAT \cite{GOAT}  (CVPR'24)                   & -        & 40.35   & \cellcolor{blue1}28.05   & 5.85    \\
DUET$^*$ \cite{chen2022duet}   (CVPR'22)                 & 36.20       & 36.28   & \cellcolor{blue1}22.58   & 3.75    \\
\rowcolor{gray!15} \ \ \ \ \textbf{+MBA (Ours)}          & 37.19   &\textbf{\underline{41.98}}   & \cellcolor{blue1}\textbf{\underline{29.57}}  & \underline{\textbf{6.05}}   \\ \midrule
\multicolumn{5}{c}{Test-unseen} 
\\ \midrule
GBE \cite{zhu2021soon}  (CVPR2021)                          &  27.88&        12.90&      \cellcolor{blue1}  9.23 &        0.45\\
GridMM \cite{wang2023gridmm}   (ICCV'23)                       & 46.20&        36.27&        \cellcolor{blue1}21.25&        4.15\\
GOAT \cite{GOAT}    (CVPR'24)                       &        -&    \underline{\textbf{40.50}} &        \cellcolor{blue1}25.18&        6.10\\
DUET$^*$ \cite{chen2022duet}   (CVPR'22)                          &        41.83&        33.44&        \cellcolor{blue1}21.42&        4.17\\
\rowcolor{gray!15} \ \ \ \ \textbf{+MBA (Ours)}     & 36.50  & 38.81  & \cellcolor{blue1}\underline{\textbf{26.16}}  & \textbf{\underline{6.28}}
\\ \bottomrule
\end{tabular}}
\label{tab:soon-sota}
\vspace{-7pt}
\end{table}

\subsection{Comparsion with SOTAs}
\label{sec: sota}

In this section, we compare the experimental results with previous state-of-the-art methods on REVERIE, R2R and SOON datasets. 
Inspired by the aforementioned analysis on the REVERIE dataset, we adopt the optimal combination $(\hat{v}^{PV}, \hat{v}^{PV})$ as our MBA method (even though it might not be optimal for R2R and SOON) to compare against SOTA methods on these three datasets.
\textbf{REVERIE}:
Table \ref{tab:reverie-sota} compares our method with previous models on the REVERIE dataset. 
We significantly boost generalizable performance, achieving SPL scores of 39.80\% and 39.53\% on the unseen validation and test sets, respectively. 
This underscores the superiority of our approach, whether navigation or object recognition.
\textbf{R2R}:  Table \ref{tab:r2r-sota} presents a comparison of results on the R2R dataset. While the navigation performance of the MBA method on the DUET baseline reaches a competitive level, the gains observed are less significant. This could be attributed to the more detailed instructions in the R2R dataset, which impose stricter constraints on the visual modality.
\textbf{SOON}:  We also conducted experiments on the SOON dataset. As shown in Table \ref{tab:soon-sota}, our method outperforms the performance of the SOTA GOAT model \cite{GOAT}, achieving significant improvements of 3.01\% and 1.01\% in SPL and RGS metrics, respectively.
Overall, evaluation results on three prominent VLN datasets demonstrate that our method not only performs comparably to, or surpasses, leading SOTA methods, highlighting its superior efficacy in enhancing generalized navigation performance across diverse vision-language navigation tasks.

\begin{table}[t]
\centering
\caption{Navigation performance on REVERIE and R2R. \textbf{Bold} highlights improvements after incorporating our MBA method.}
\begin{tabular}{l|cccc}
\toprule
 \multicolumn{5}{c}{REVERIE Val-unseen} \\
 Methods   & SR     & SPL   & RGS   & RGSPL  \\ \midrule
DSRG \cite{wang2023dsrg}    \ \ \           & 47.83  & 34.02 & 32.69 & 23.37          \\
\rowcolor{gray!20} \ \ \ \ \textbf{+MBA(Ours)} \ \ \  & \textbf{50.64}& \textbf{34.55}& \textbf{34.37}&23.32  \\
LAD \cite{li2023lad}   \ \ \                   & 57.00  & 37.92 & 37.80  & 24.56 \\
\rowcolor{gray!20} \ \ \ \ \textbf{+MBA(Ours)} \ \ \  & 56.64& \textbf{39.74}& 37.21& \textbf{26.00}  \\
DUET \cite{chen2022duet}              & 51.55  & 35.75 & 35.19 & 24.40 \\
\rowcolor{gray!20} \ \ \ \ \textbf{+MBA (Ours)}     \ \ \         & \textbf{52.29}  & \textbf{39.80} & \textbf{35.81} & \textbf{27.44}  \\ \midrule

\multicolumn{5}{c}{R2R Val-unseen} \\
             Methods           & TL$\downarrow$     & SR     & SPL   & NE$\downarrow$   \\ \midrule
DSRG  \cite{wang2023dsrg}      & -      & 73     & 61    & 3.00\\
\rowcolor{gray!20} \ \ \ \ \textbf{+MBA(Ours)}   \ \ \     & 13.46  & 72.54  & \textbf{61.95} & 3.07 \\
DUET \cite{chen2022duet}               & 12.56  & 71.99  & 63.18 & 3.17 \\
\rowcolor{gray!20} \ \ \ \ \textbf{+MBA(Ours)}    \ \ \    & 13.39  & \textbf{72.71}  & \textbf{63.19} & \textbf{3.05} \\ \bottomrule
\end{tabular}
\label{tab:agnostic}
\end{table}


\subsection{Architecture Agnostic}

Another merit of the proposed MBA framework is agnostic to topology-based VLN agents, which allows for seamless integration with various existing VLN agents in a plug-and-play manner, without meticulously modifying the model architecture or introducing more environment annotations. As demonstrated in Table \ref{tab:agnostic}, we also employ  DSRG \cite{wang2023dsrg}, LAD as baselines. Concretely, we achieved an impressive 1.82\% SPL improvement for LAD on the REVERIE dataset. Similarly, we attained a 0.95\% enhancement in the SPL metric for DSRG on the R2R dataset. Notably, DSRG still utilizes ViT visual features.
When compared to these baselines, our approach consistently improves the SPL metric across both the REVERIE and R2R datasets, suggesting that increasing the number of branches helps agents better balance the success rate and navigation length. Moreover, whether using the ViT-based DSRG or the CLIP-based DUET, combining them with our approach yields limited improvements on the R2R dataset. This may be because the performance metrics on R2R are already near optimal, making further gains challenging.

\begin{figure}
    \centering
    \includegraphics[width=0.95\linewidth]{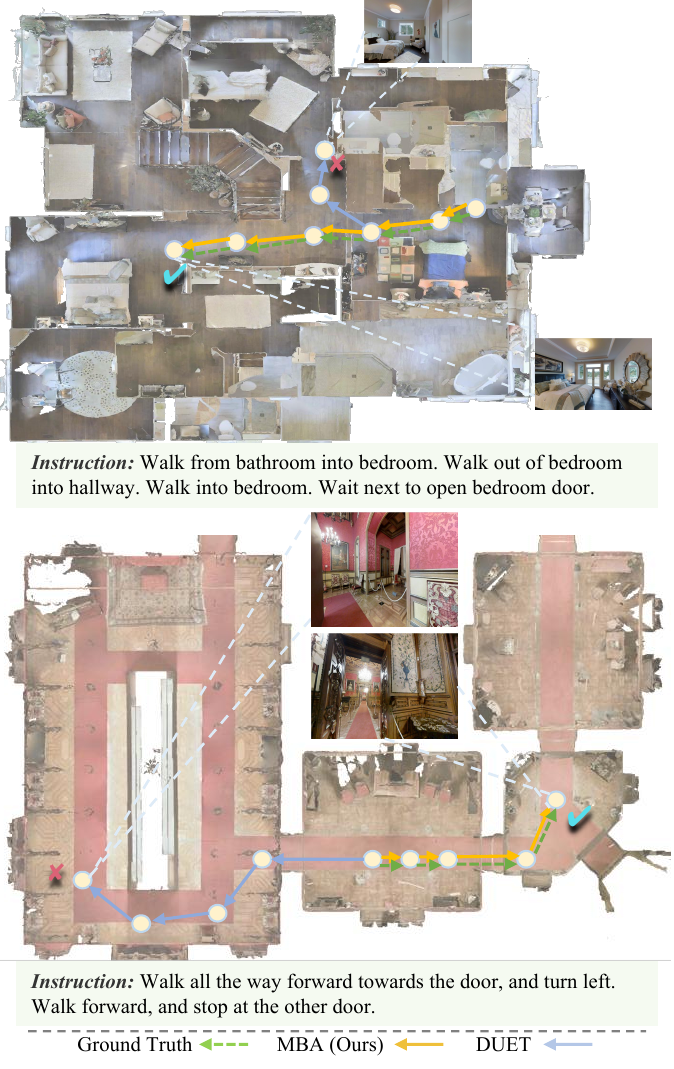}
    \caption{Predicted trajectories of MBA and DEUT on R2R Val-unseen split.}
    \label{fig:small}
\vspace{-10pt}
\end{figure}

\subsection{Qualitative Analysis }

As shown in Fig.~\ref{fig:small}, we visualize the top-down navigation trajectories on the R2R validation unseen split. In the first example, our MBA agent demonstrates superior performance by accurately stopping near the ``bedroom door." In the second example, our method effectively follows the instructions, achieving correct navigation. In contrast, the DUET agent, which relies solely on RGB-based environmental representations, tends to exhibit erratic behavior, compromising its ability to follow instructions accurately and reach the target location. This highlights the potential of augmenting the agent branch with necessary visual perturbations to robustly improve the navigation performance in unseen environments.

\section{Conclusion} 


This study introduces three additional visual perturbations as inputs to our proposed multi-branch architecture to explore the potential effect of diverse visual information. 
We experimentally verify that topology-based agents can benefit from diverse visual inputs beyond standard RGB images, even random noise can significantly enhance the navigation generalization performance, which offer a fresh perspective into VLN.
Based on these surprise experimental findings, our approach not only outperforms existing methods on three VLN benchmarks but also highlights the importance of visual perturbation in improving generalization.  Nevertheless, the underlying mechanisms driving these improvements require further reconsideration.

 \section{Limitation and Future Work}

In this work, we demonstrated that incorporating additional branches with necessary visual perturbations can enhance performance without modifying the base architecture. However, several limitations warrant further investigation:
(1) Although our proposed method yields performance improvements through the addition of branches and the strategic application of noise, future research should focus on elucidating the underlying mechanisms responsible for these navigation performance gains.
(2) Building upon existing methodologies~\cite{Li_2024,zhang2024beyondrgb,he2024fda}, we will explore the integration of complementary environmental representations, such as textual semantic knowledge, depth information, and frequency features. This approach aims to inject an inductive bias that encourages the model to develop a more comprehensive understanding of, and grounding in, the environment.






\end{document}